\documentclass[letterpaper, 10 pt, conference]{ieeeconf}
\IEEEoverridecommandlockouts
\usepackage{times}
\usepackage{graphics}
\usepackage{graphicx}
\usepackage{subfigure}
\usepackage{amsmath,amssymb,amsopn,amstext,amsfonts}
\usepackage{cancel}
\usepackage[space]{cite}
\usepackage{pdfsync}
\usepackage{balance}
\usepackage{color}
\usepackage{mathtools}
\usepackage{bm}

\usepackage{diagbox}
\usepackage{float}
\usepackage{epstopdf}
\usepackage{pifont}
\usepackage{fixltx2e}
\usepackage{amsmath}
\usepackage{multirow}
\usepackage{url}
\usepackage{verbatim}
\usepackage{caption}
\usepackage{adjustbox}
\usepackage{booktabs}
\usepackage{threeparttable}
\usepackage{makecell}
\usepackage{tabularx}
\usepackage{arydshln}
\usepackage[normalem]{ulem}
\usepackage{algorithm}
\usepackage{ifthen}
\usepackage{algorithmic}
\usepackage{array}
\usepackage{textcomp}
\usepackage{stfloats}

\makeatletter
\let\NAT@parse\undefined
\makeatother
\usepackage[linkcolor=red,citecolor=green,urlcolor=red,colorlinks=true]{hyperref}

\title{\LARGE \bf ODD-SEC: Onboard Drone Detection with a Spinning Event Camera}
\author{Kuan Dai$^{\ast}$, Hongxin Zhang$^{\ast}$, Sheng Zhong, Yi Zhou% <-this % stops a space
\thanks{All authors are with the Neuromorphic Automation and Intelligence Lab (NAIL) at School of Artificial Intelligence and Robotics, Hunan University, Changsha, China. }
\thanks{$\ast$ denotes equal contribution.}
\thanks{Corresponding author: Yi Zhou. Email: {\tt\small eeyzhou@hnu.edu.cn}.}
\thanks{This work was supported by the National Key Research and Development Project of China under Grant 2023YFB4706600.
}
}

\definecolor{myteal}{RGB}{0,128,128}
\begin{document}
\maketitle
\thispagestyle{empty}
\pagestyle{empty}
\begin{abstract}
The rapid proliferation of drones requires balancing innovation with regulation. 
To address security and privacy concerns, techniques for drone detection have attracted significant attention.
Passive solutions, such as frame camera-based systems, offer versatility and energy efficiency under typical conditions but are fundamentally constrained by their operational principles in scenarios involving fast-moving targets or adverse illumination.
Inspired by biological vision, event cameras asynchronously detect per-pixel brightness changes, offering high dynamic range and microsecond-level responsiveness that make them uniquely suited for drone detection in conditions beyond the reach of conventional frame-based cameras.
However, the design of most existing event-based solutions assumes a static camera, greatly limiting their applicability to moving carriers—such as quadrupedal robots or unmanned ground vehicles—during field operations.
In this paper, we introduce a real-time drone detection system designed for deployment on moving carriers. 
The system utilizes a spinning event-based camera, providing a $360^\circ$ horizontal field of view and enabling bearing estimation of detected drones. 
A key contribution is a novel image-like event representation that operates without motion compensation, coupled with a lightweight neural network architecture for efficient spatiotemporal learning. 
Implemented on an onboard Jetson Orin NX, the system can operate in real time. 
Outdoor experimental results validate reliable detection with a mean angular error below $2^\circ$ under challenging conditions, underscoring its suitability for real-world surveillance applications. 
We will open-source our complete pipeline to support future research.
\end{abstract}

% TODO: In this sentence "Outdoor experimental results validate...", add some numbers to justify how much your method outperform existing method!
% Outdoor experimental results validate reliable detection with a mean angular error below $2^\circ$, outperforming frame-based baselines under fast motion and adverse illumination, underscoring its suitability for real-world surveillance applications. 
% 
% \input{chapters/Multimedia_material}
\section{Introduction}
\label{sec:introduction}

Drones are increasingly adopted in fields such as aerial photography, precision agriculture, logistics, search and rescue, and environmental monitoring~\cite{siean2021taking,hong2023logistics,sowmya2024creating}, owing to their low cost and versatility. However, improper or malicious use poses risks to safety, security, privacy, and airspace operations~\cite{rahmani2024working}.
Addressing these issues is therefore crucial to ensure the safe and beneficial integration of drones into society.

\begin{figure}[t]
    \centering
    \includegraphics[width=1\linewidth]{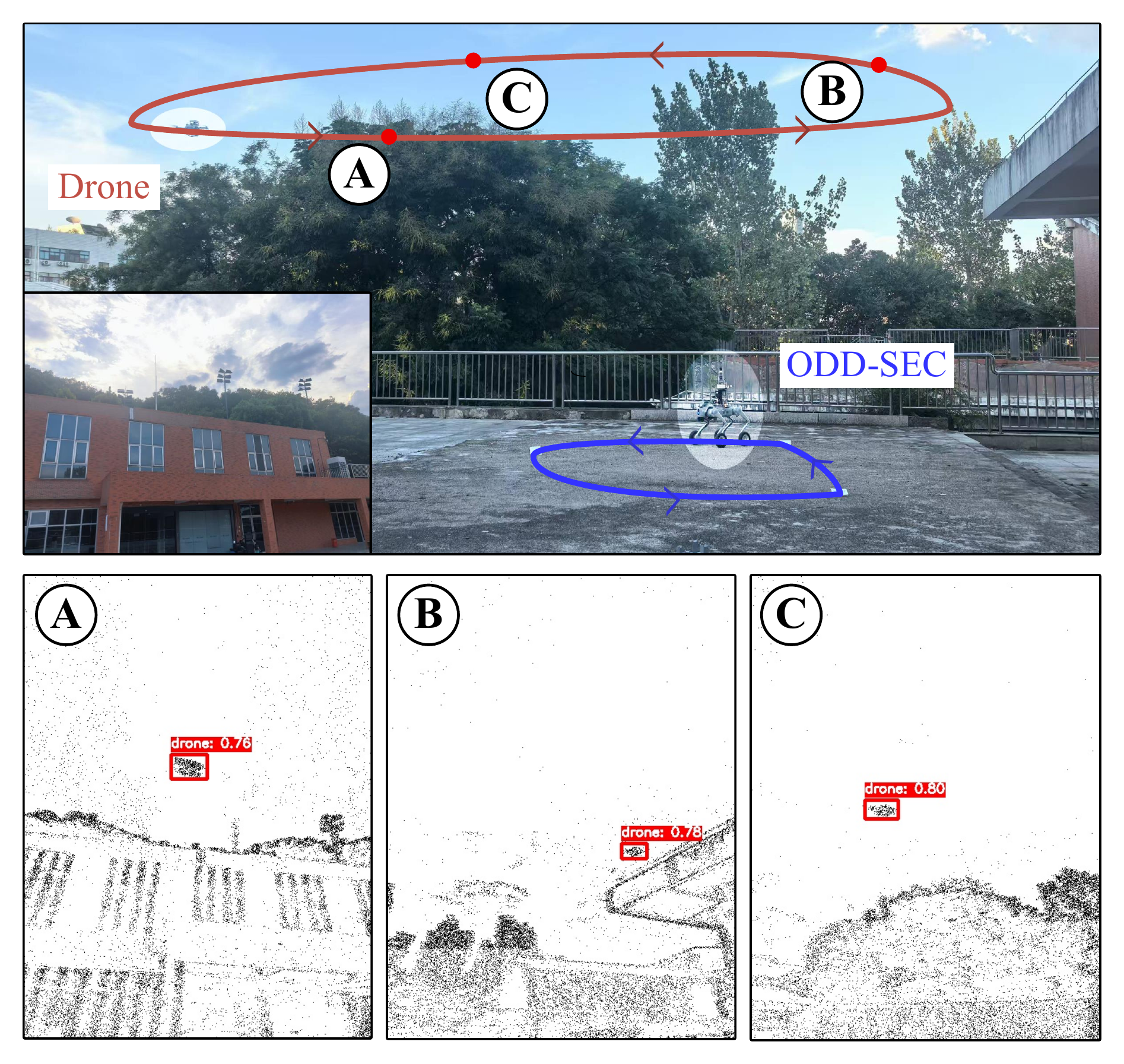}
    \caption{
    \textit{Illustration of our system operating in an outdoor scenario.} 
    Top: The scene with the trajectories of the drone (in red) and our device (in blue), with an inset at the bottom-left showing the rear view of the same scene.
    Bottom: The detection results from our system.
    Specific measurement locations are marked with circled letters.}
    \label{fig: eye_catcher}
    \vspace{-0.0em}
\end{figure}

% 我们的系统在实景下工作的示意图。上部：实景图以及运动的轨迹 蓝色的轨迹是无人机的运动 红色的轨迹是我们的设备运动的轨迹  下部：相机检测的结果

% TOP的两句合并(in red) 标号的对应说明
% Inspired by the human visual system, event-based cameras asynchronously report per-pixel brightness changes~\cite{gallego2020event}, offering microsecond temporal resolution and high dynamic range (HDR) for robust drone detection beyond conventional cameras.
Inspired by the human visual system, event-based cameras asynchronously report per-pixel brightness changes~\cite{gallego2020event}, providing rich motion information, high dynamic range (HDR), and fine-grained temporal cues embedded in the event stream.
Zhou \textit{et al.}~\cite{zhou2021event} leverage these unique characteristics to accurately identify independently moving objects from the fine-grained asynchronous event stream, while Wang \textit{et al.}~\cite{wang2024asynchronous} propose AEB Tracker, achieving over 50 kHz update rates under challenging illumination.
However, the inherently limited field of view (FoV) of event cameras restricts their detection range, as the unpredictable flight trajectories of drones can easily cause them to leave the camera’s observable range, resulting in detection failure.
% \joey{TODO: Did we compare against mentioned methods and demonstrate advantages in terms of long-term detection??}
% \dai{I've added it in the following paragraphs.}

% To address this limitation, several researchers have proposed approaches to expand the effective coverage of event-based vision systems.
% Researchers have leveraged event cameras' unique properties to circumvent the FoV limit, developing novel approaches for expanding coverage.
% Chaumette \textit{et al.}\cite{chaumette2008visual} proposed an active visual servoing framework to dynamically control the camera's motion.
% Vasseur \textit{et al.}\cite{vasseur2023omnidirectional} integrated a prism-mirror assembly, extending detection coverage while reducing alignment complexity.
% Da \textit{et al.} ~\cite{da2025new} eliminated mechanical complexity by deploying two back-to-back fisheye lenses, achieving full 360° omnidirectional perception without moving parts.
A recent effort to address this limitation is the SFERA system proposed by Da \textit{et al.}~\cite{da2025new}, which employs back-to-back event cameras to achieve full $360^\circ$ omnidirectional coverage, enabling drone detection and angular localization relative to the device.
% However, the system was validated only on stationary platforms and occasionally suffers from unstable outputs, such as spurious peaks caused by tracker re-initialization when confidence drops below a threshold.
% These limitations constrain its applicability in dynamic, real-world scenarios.}
% These limitations significantly constrain its real-world applicability, particularly in dynamic deployment scenarios such as on moving platforms.
This approach, however, implicitly assumes a static viewpoint, as all validation is conducted on a tripod-mounted system. 
Such a design is insufficient for dynamic deployment scenarios such as on moving carriers, where unaddressed ego-motion could compound known system instabilities, e.g., tracker failures requiring re-initialization, and corrupt the entire perceptual pipeline.
% While these solutions achieve effective panoramic target detection, platform mobility becomes essential for comprehensive scene monitoring. 
% However, such movement introduces complex kinematic uncertainties, and ego-motion induces events from static background objects, making it challenging for fisheye-based systems to detect target objects.

% To overcome these challenges, we propose ERDS (Event-based Rotating Detection System), consisting of an event camera, an edge computer, and a rotating module. 
To overcome these challenges, we propose ODD-SEC: an onboard system that integrates a spinning event camera, a spinning module, and an edge computer for real-time drone detection.
The spinning camera enables omnidirectional drone detection with panoramic horizontal coverage, overcoming FoV limitations and supporting long-term tracking.
We integrate all modules into a self-contained system capable of deployment on moving carriers, such as a quadruped robot, for effective drone detection, as shown in Fig. \ref{fig: eye_catcher}.
The core contributions of this paper are as follows:

\begin{itemize} 
    \item A novel event-based drone detection system design enabling 360° panoramic perception via a platform-mounted spinning event camera.
    This design effectively expands the FoV, overcoming the inherent FoV limitations of traditional event camera setups.
    \item A temporally-aware variant of the YOLOX network, taking a specific image-like representation of events as input, effectively addresses the challenges of massive data volume and motion blur caused by camera ego-motion and carrier movement. 
    % The system achieves real-time drone position detection on Jetson Orin NX, demonstrating practical deployment capability.
    \item A comprehensive outdoor experiment conducted on a moving carrier validates the system's capability to achieve reliable detection performance.
    The system achieves real-time drone detection and bearing estimation on Jetson Orin NX, demonstrating practical deployment capability.
    We will release our pipeline as an open-source code for future research in this field.
\end{itemize}
% TODO position 分开
% The rotating platform is additionally equipped with a localization module.
% With the detected drone position in the image plane, we can derive its angle relative to the rotating module.

% The implemented system achieves autonomous 30-FPS detection on the embedded Jetson Orin NX platform.

\section{Related Work}

% 事件相机的无人机检测以及全景事件相机发展现状：固定平台检测方式（存在限制）—>全景的
% 事件识别模式
% Like its standard-vision counterparts, event-based VO/SLAM can be broadly divided into model-based methods and learning-based methods.
% The former estimates the camera trajectory by processing events and explicitly constructing photometric and geometric constraints, while the latter utilizes deep networks to learn implicit information from event streams or event representations for trajectory estimation.
In line with the focus of our work, we review two closely related areas: event-camera-based object detection, which underpins our detection module, and FoV extension methods, which are essential to overcoming the inherent FoV limitations of event cameras.

\subsection{Object Detection for Event Cameras}

Event-based object detection methods can be broadly categorized into two paradigms according to their representation of events: point-based models and image-based models.
The former paradigm leverages the sparse and asynchronous nature of event data, often employing spiking neural networks (SNNs)~\cite{spikeyolo, sanyal2024EV-Planner, emsyolo} or graph-based processing techniques~\cite{aegnn, nvs}. 
While effectively preserving the spatiotemporal distribution of raw events, these methods often face challenges related to data sparsity and computational efficiency.
In contrast, the image-based models convert event streams into dense, frame-like representations, enabling the use of well-established convolutional neural networks.
Some works in this line, such as~\cite{GET-Net, GWD-Net, RecurrentTransformerNet, DE_sodformer_2023}, extract spatiotemporal features using stacked representations from consecutive short time windows.
Although these approaches may not fully exploit the temporal information inherent in event streams, they have demonstrated competitive detection performance and remain dominant in practical applications.

Beyond these primary categories, several studies explore hybrid systems that combine event cameras with conventional frame-based sensors~\cite{tomy2022, chen2025eMoE-Tracker, Magrini2025}, leveraging complementary attributes of both modalities to enhance detection robustness.
Furthermore, for specific moving objects such as drones, some researchers have analyzed the kinematic models of targets and achieved detection of specific objects by leveraging such models, without the need for training deep neural networks~\cite{sanket2021evpropnet, zhang2025EvDetMAV}.

\subsection{FoV Extension Methods}

% FoV extension
% 分类：

% 机械结构

% 鱼眼 fisheye Panoramic Annular Lens

% 拓展FoV的方法有多种，可以分为利用机械结构使相机移动或使用特殊镜头
% 机械结构有
% visual servoing 或 using a rotating device

% 特殊镜头有fisheye和 Panoramic Annular Lens等

% Methods for expanding the FoV can generally be categorized into two groups.
% The first group leverages mechanical structures to physically move the camera, thereby enlarging the observable area. Representative approaches include visual servoing for active camera motion\cite{saxena2017exploring,bateux2018training} and the use of rotating devices to achieve panoramic coverage\cite{schraml2015event}.
% The second group relies on optical designs, where specialized lenses are employed to inherently capture a wider FoV, such as fisheye lenses~\cite{da2025new,xie2024omnidirectional} and Panoramic Annular Lenses (PALs)\cite{wang2022lf}.

The most straightforward approach to expanding a camera’s FoV is through specialized lenses, such as fisheye lenses~\cite{da2025new,jaisawal2024airfisheye}. 
Another commonly used option is Panoramic Annular Lenses; for instance, the lens adopted in~\cite{wang2022lf} provides a FoV of $360^\circ \times (40^\circ$–$120^\circ)$, directly enabling panoramic coverage. 
However, since the camera resolution remains unchanged, the resulting images tend to be more blurred.

An alternative strategy involves extending the FoV through camera motion.
For example, visual servoing techniques~\cite{saxena2017exploring,bateux2018training} enable active control of camera pose to dynamically expand the effective FoV, allowing the system to track targets and cover larger workspaces—though at the cost of increased system complexity due to perception–control coupling. 
A simpler motion-based approach is to employ continuous camera rotation. 
Schraml \textit{et al.}~\cite{schraml2015event} develop an event-driven stereo vision system using a pair of rotating single-column Dynamic Vision Sensors, achieving real-time $360^\circ$ 3D panoramic reconstruction with high dynamic range performance.

\begin{figure}[t]
    \vspace{1.0em}
    \centering
    \includegraphics[width=0.8\linewidth]{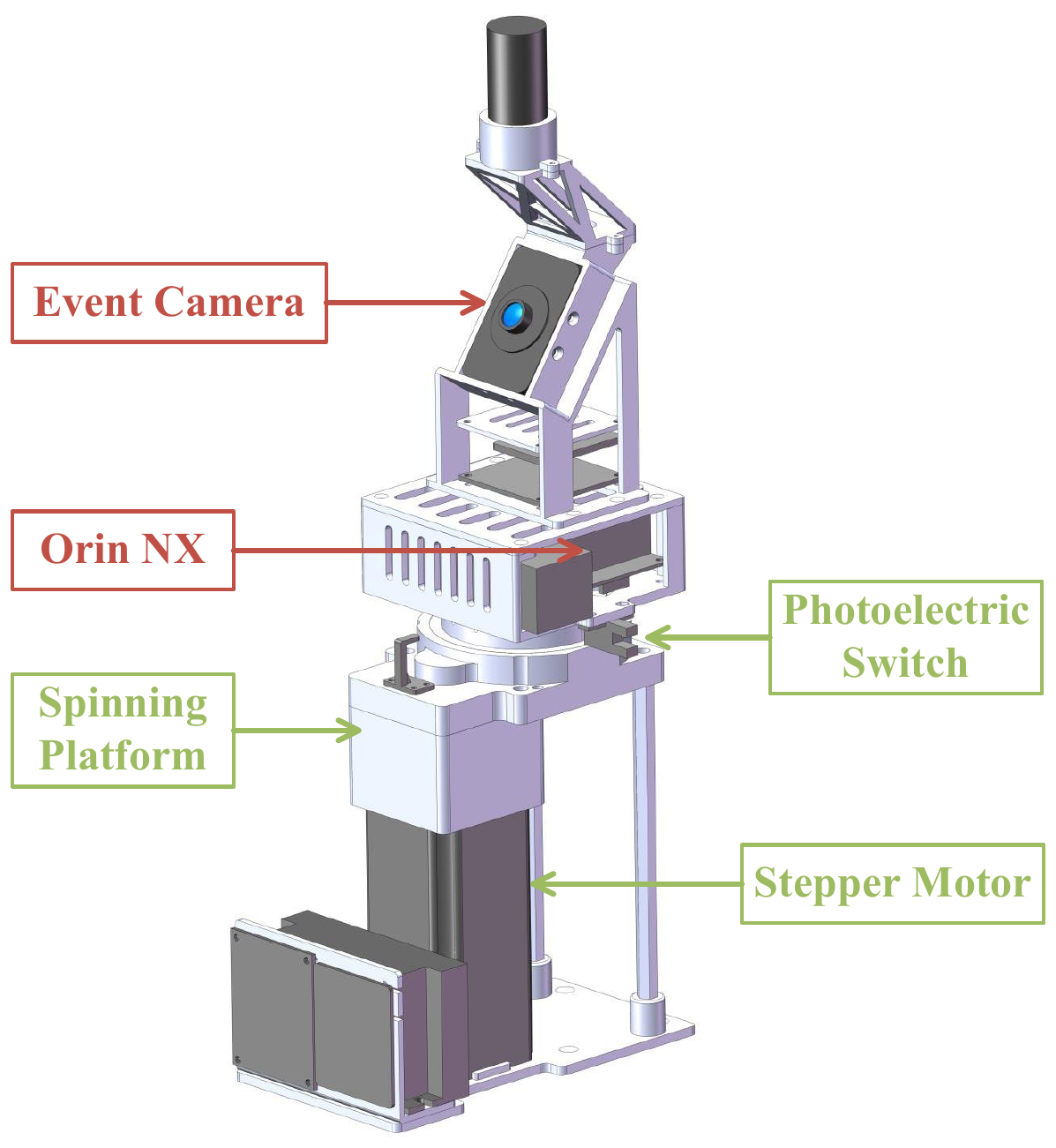}
    \caption{
    \textit{CAD model of the system.} 
    % Green labels denote devices of the actuation module, and red labels denote devices of the perception–computing module. 
    Red and green boxes represent the actuation module and the perception-computing module, respectively.}
    \label{fig: hardware_module}
    \vspace{-2.0em}
\end{figure}

\section{Methodology}
\label{sec:methods}

In this section, we detail the method of our system. 
We first introduce the hardware setup and mounting details of the proposed system (Sec.~\ref{subsec:hardware_setup}).
Then, we discuss the algorithms used in our system (Sec.~\ref{subsec:algorithm_design}), including the drone detection network and positioning strategy.
%Edit: Second, we describe the proposed network and how we compute bearing vector from the network outputs(Sec.~\ref{subsec:algorithm_design}).
% 这一节我们介绍系统 . First, we 介绍了硬件配置和安装方式.Second, 我们介绍了主要的算法,包括Network Design和bearing Estimation(Sec.~\ref{subsec:algorithm_design})

\subsection{Hardware Setup}
\label{subsec:hardware_setup}

The CAD model of our system hardware is shown in Fig.~\ref{fig: hardware_module}, with the specifications of the primary devices listed in Tab.~\ref{tab: hardware_list}.
The hardware consists of two modules: an actuation module responsible for spinning the camera, and a perception-computing module dedicated to data recording and processing.

\begin{table}[t]
\setlength{\tabcolsep}{14pt}
\vspace{1.0em}
\centering
\caption{Hardware setup and parameters.}
\resizebox{0.42\textwidth}{!}{%
\begin{tabular}{p{0.12\textwidth} p{0.18\textwidth}}
    \toprule
      \makecell*[c]{\textbf{Devices}}  & {\textbf{Parameters}} \\
     \midrule
     \makecell*[c]{Event Camera\\(DVXplorer)} & \makecell[l]{640$\times$480 1/2.5'' \\ FoV: 77.32°$\times$ 61.93°\\ 3.6-mm lens}\\
     \midrule
     \makecell*[c]{ Stepper Motor } &  
     \makecell[l]{ Step angle: 1.8° \\ Torque: 2.8 $ N \cdot m $ \\ Controller: TB6600 }\\
     \midrule
     \makecell*[c]{Spinning Platform} & \makecell[l]{Reduction ratio: 5:1}\\
     \midrule
     \makecell*[c]{Photoelectric \\ Switch} & \makecell[l]{Response time: $<$0.5 ms } \\
     \bottomrule 
\end{tabular}
}
\label{tab: hardware_list}
\vspace{-0.5em}
\end{table}

% 由两个组成consists of 感知与运算  and :
% 1包含...2包含... 捕捉信息/相机旋转

% 两个模块反过来讲:1.执行模块旋转及目的 2.感知计算 事件相机 相机选用原因删掉 

% 表格引用放在图的后面 表格表头加粗 标题移到上面
% 图改一下 两个色

% The hardware system consists of two parts: the sensing-computing module, responsible for data acquisition and processing, and the actuation platform, which integrates a motor-dirven rotary platform for controlled rotation.
% The setup can be divided into two parts: the first consists of the sensors and the computing platform, and the second is the rotating platform, which includes the motor and rotary platform. 
% The CAD model of the whole system is shown in Fig.~\ref{fig: hardware_module}.

% Spinning Stage
% Spinning Mechanism
% Spin Module
% Spinning Mount
% Spinning Assembly
% Spinning Unit
% Spinning Structure

% The sensing-computing module is mounted on a rotating rotary platform driven by a stepper motor.
To achieve a $360^\circ$ horizontal FoV, the actuation module spins the sensing–computing module via a stepper-motor-driven spinning platform.
The motor is controlled by a battery-powered STM32-based driver for precise speed and direction regulation.
During startup, the motor initially runs at a lower speed and then gradually accelerates to a constant angular velocity of 0.95 rad/s (to be distinguished from the Pulses Per Second (PPS) rate; detailed rationale in Sec.~\ref{subsec: experimental_details}), thereby avoiding excessive acceleration that can lead to step loss. 
A photoelectric switch mounted on the spinning platform defines the zero-point direction.
It is triggered once per rotation and provides a synchronization signal to the event camera, ensuring angular alignment and accurate orientation recovery.

% A photoelectric switch is connected to the event camera. 
% Whenever the rotating platform reaches the position of the blocking plate, the switch sends a trigger signal to the camera, enabling accurate detection of the zero-point position in each rotation cycle.
\begin{figure}[t]
    \centering
    \includegraphics[width=1\linewidth]{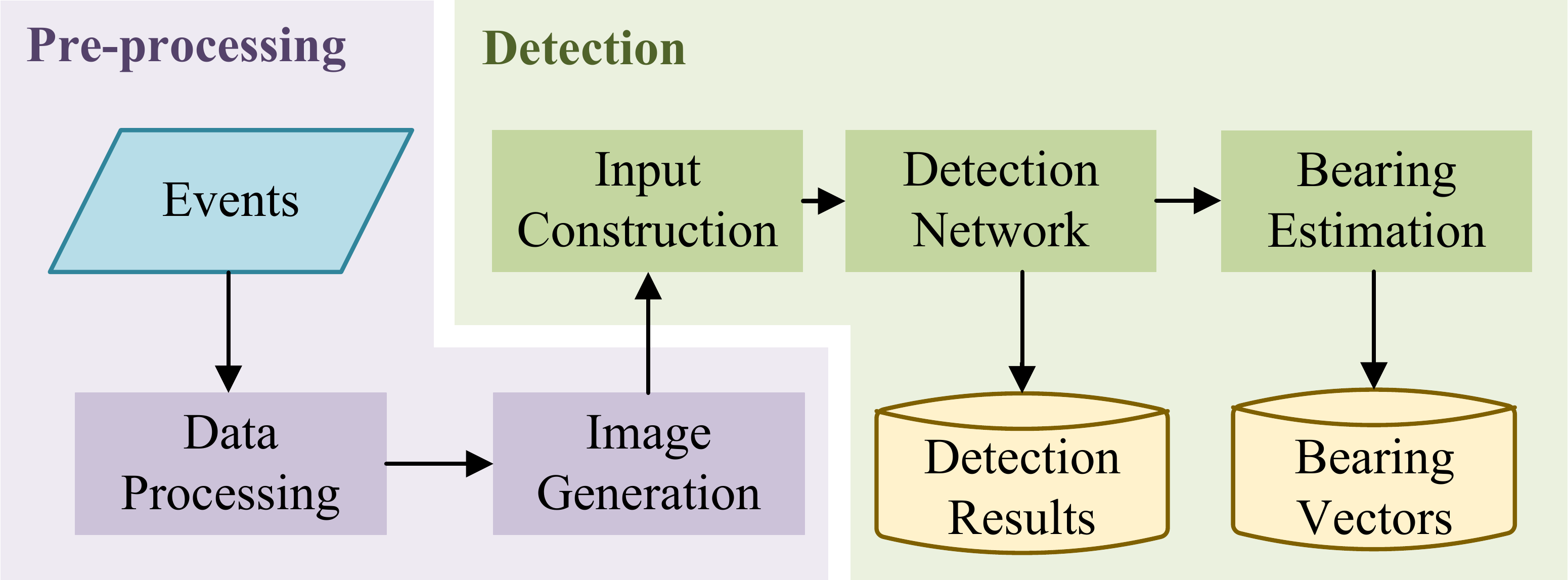}
    \caption{
    \textit{Flowchart of the proposed algorithm.} 
    The two modules are executed independently.
    The Pre-processing part receives data from the event camera and converts it into image slices.
    The Detection part receives and processes the image stream, performs network inference, and calculates the relative bearing to the system based on the detection results.}
    \label{fig: pipeline}
    \vspace{-2.0 em}
\end{figure}

The DVXplorer event camera serves as the primary sensor.
To maximize coverage of potential drone flight areas, the camera is oriented vertically with a 35° upward tilt.
The spinning platform’s rotation axis passes through the camera’s optical center. 
% To address potential massive data transmission challenges caused by continuous rotation, an edge computing unit is co-rotated with the camera assembly.
To address the potential high-bandwidth data transmission challenges caused by continuous rotation, the edge computing unit spins together with the camera assembly.
This integrated computing unit directly receives raw sensor data and executes the proposed algorithm, achieving end-to-end real-time drone detection without relying on slip rings or wireless transmission for high-bandwidth data.

% After being rotated 90° counterclockwise such that its width aligns with the vertical axis, the camera is further mounted on the platform at an inclination angle of 35° relative to the horizontal plane, in order to optimize the field of view.
% A photoelectric switch is connected to the event camera. 
% Whenever the rotating platform reaches the position of the blocking plate, the switch sends a trigger signal to the camera, enabling accurate detection of the zero-point position in each rotation cycle.
% The data processing unit consists of an NVIDIA Orin NX embedded computer running Ubuntu 20.04 and ROS1 Noetic, offering sufficient computational resources for real-time execution of complex algorithms.

All devices are battery-powered, enabling the system to operate independently.
In addition, the actuation module is equipped with a mounting interface, facilitating integration with mobile carriers.

\subsection{Algorithm Design}
\label{subsec:algorithm_design}
The proposed algorithm (shown in Fig.~\ref{fig: pipeline}) takes in event data output by the spinning event camera, detects the 2D position of the drone, and calculates the bearing vector between the drone and our system.
In the pre-processing module, raw events are transformed into a multi-slice spatio-temporal representation (MSR) to decouple target kinematics from rotation-induced spatial smearing (Sec.~\ref{sec:Repre}).
In the subsequent detection module, these slices are then processed by the Temporal Fusion Module (TFM), embedded within the YOLOX framework~\cite{ge2021yolox}, to align robust features and adaptively suppress background noise (Sec.~\ref{sec:tfm}).
For precise localization under extreme target scale variations, we employ a scale-aware loss that explicitly decouples the penalty terms for central point distance and aspect ratio (Sec.~\ref{sec:loss}).
Finally, the bearing estimation module translates these 2D detections into precise relative angles using camera intrinsics and real-time orientation data (Sec.~\ref{sec:bearing_estimation}).

% To address this, our pipeline transforms raw events into a multi-slice spatio-temporal representation (MSR) to decouple target kinematics from rotation-induced spatial smearing (Sec.~\ref{sec:Repre}).
% Slices are then processed by the Temporal Fusion Module (TFM) to align robust features and adaptively suppress background noise (Sec.~\ref{sec:tfm}).
% For precise localization under extreme target scale variations, we employ a scale-aware loss formulation (Sec.~\ref{sec:loss}).
% Finally, the bearing estimation module translates these 2D detections into precise relative angles using camera intrinsics and real-time orientation data (Sec.~\ref{sec:bearing_estimation}).
% And based on the detection outputs, the bearing estimation module (see Sec.~\ref{sec:bearing_estimation}) calculates the relative angle using the detected 2D position, camera intrinsics, and real-time camera orientation, outputting a bearing vector applicable to downstream tasks.

\begin{figure}[t]
    \vspace{1.0em}
    \centering
    \includegraphics[width=1\linewidth]{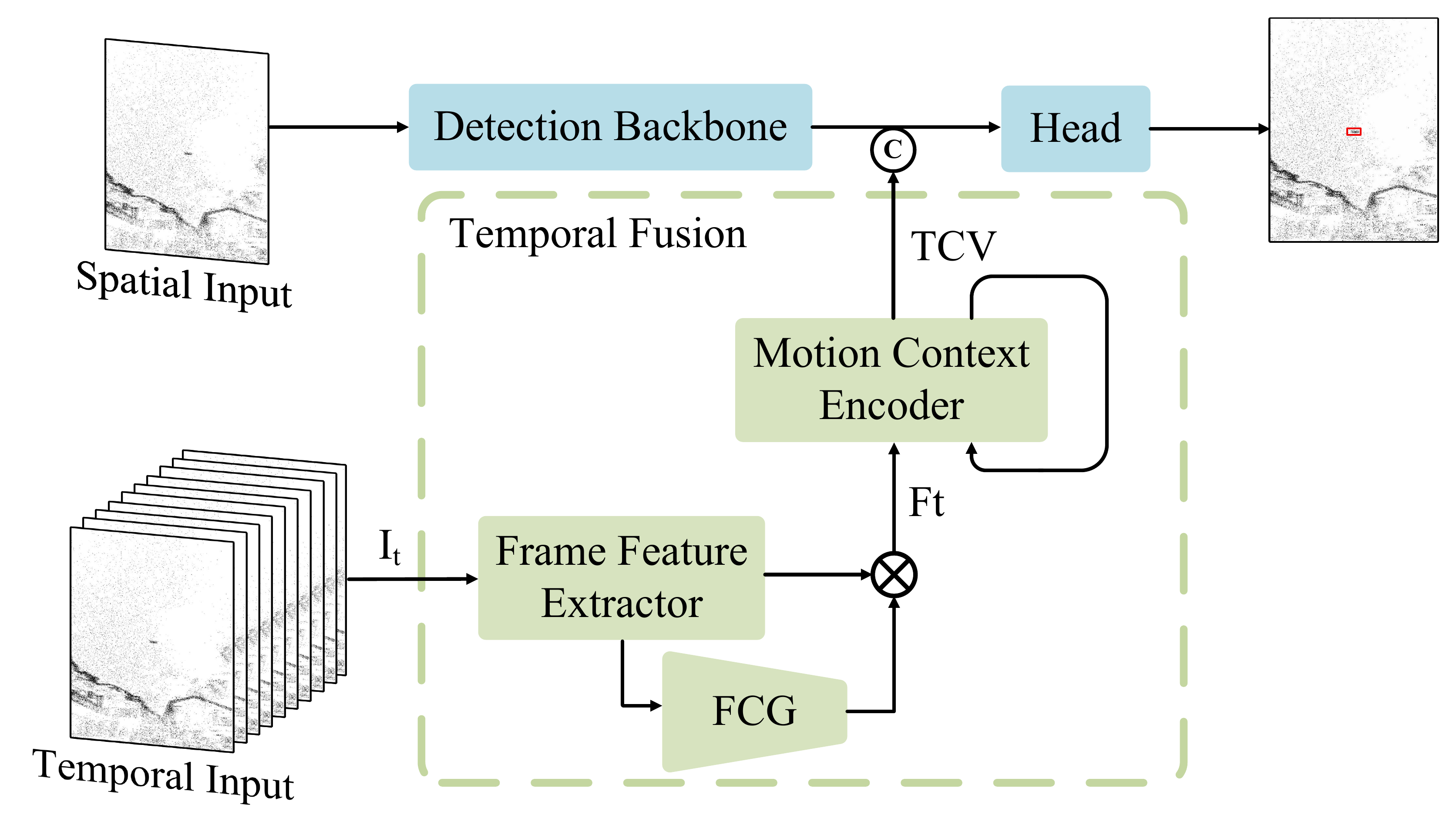}
    \caption{
    \textit{Overview of the proposed detection network.} 
    Our novel Temporal Fusion Module (highlighted in a green dashed border) enhances detection by analyzing a sequence of frames that capture motion relationships. 
    It employs Feature Channel Gating (FCG) to adaptively weight features, aggregates Temporal Context Vector (TCV) across the segment, and fuses it with features from the detection backbone to improve the network’s target recognition accuracy.}
    \label{fig: overview_network}
    \vspace{-1.5 em}
\end{figure}

\subsubsection{Multi-slice Spatio-temporal Representation}
\label{sec:Repre}

Conventional 2D event representations (e.g., Time Surface~\cite{HOTS2017}) suffer from severe information loss during continuous rotation due to dense event overwriting. 
To preserve fine-grained temporal cues and mitigate rotation-induced spatial smearing, we construct a sequential representation by dividing a fixed temporal window $\Delta T$ into $N$ uniform sub-intervals.
The $i$-th spatio-temporal slice $S_i$ is generated by:

\begin{equation}
    \begin{aligned}
        S_i(x, y) = \sum_{e_k \in \Delta t_i} p_k \cdot \delta(x - x_k, y - y_k),
    \end{aligned}
\end{equation}
where $e_k = (x_k, y_k, t_k, p_k)$ represents an event triggered at timestamp $t_k$ with polarity $p_k$ at spatial coordinates $(x_k, y_k)$, 
$\Delta t_i = [T + \frac{i-1}{N}\Delta T, T + \frac{i}{N}\Delta T]$ is the $i$-th consecutive interval, and $\delta(\cdot)$ denotes the Dirac delta function.
By partitioning the accumulation window into multiple slices, this representation effectively captures motion patterns while mitigating ego-motion-induced blur.

\subsubsection{Temporal Fusion Module}
\label{sec:tfm}

% The multi-slice sequence is processed by a dual-branch network architecture, as shown in Fig.~\ref{fig: overview_network}.
% To extract stable target representations from the dynamic background noise, we introduce a lightweight Temporal Fusion Module (TFM) integrated into the neck of the detector.
% The core component of the TFM is the Feature Channel Gating (FCG) mechanism, which acts as an adaptive filter to suppress background flow and enhance salient target signals.
% Given the feature map $F_t$ extracted from the current slice, the FCG refinement is formulated as:

Processing the generated slices as independent frames fails to capture the temporal evolution of the scene, limiting the network's ability to utilize crucial motion cues.
To capture the scene's temporal evolution, we propose a dual-branch architecture, as illustrated in Fig.~\ref{fig: overview_network}.
The primary backbone extracts spatial features from the current slice, while TFM processes the entire MSR to encode motion context.
While the primary backbone extracts spatial features from the current slice, MSR is processed by TFM to encode a motion context.
Internally, the TFM first employs a Feature Channel Gating (FCG) mechanism to explicitly evaluate the informational value of each feature channel.
Specifically, the FCG utilizes global average pooling to aggregate spatial information and capture global channel statistics. 
These statistics are then used to generate a dynamic weighting vector that recalibrates the channel responses.
By applying these learned weights, the FCG explicitly amplifies salient target features and suppresses redundant background noise induced by continuous ego-motion.
The gating process yields a refined intermediate feature $F_t$, which is formulated as:
\begin{equation}
    \begin{aligned}
        F_t = \sigma(\mathbf{W}_2 \cdot \text{ReLU}(\mathbf{W}_1 \cdot \text{GAP}(F_{ext}))) \otimes F_{ext},
    \end{aligned}
\end{equation}
where $F_{ext}$ represents the features from the frame feature extractor, $\text{GAP}(\cdot)$ denotes global average pooling, $\sigma(\cdot)$ is the sigmoid activation function, $\mathbf{W}_1$ and $\mathbf{W}_2$ are learnable weight matrices, and $\otimes$ represents channel-wise multiplication.

The enhanced feature $F_t$ is subsequently processed by a Motion Context Encoder to capture cross-slice kinematics, yielding the Temporal Context Vector (TCV). 
This vector is concatenated with the appearance features from the spatial anchor frame. 
This architectural design enables the network to implicitly learn spatial alignment and spatio-temporal correlations between consecutive slices, bypassing the need for computationally prohibitive optical flow estimation or precise IMU-based ego-motion compensation.

\subsubsection{Scale-Aware Loss}
\label{sec:loss}

The YOLOX framework originally employs standard IoU loss, but this metric is suboptimal for targets with extreme scale variations.
Specifically, the changing distance between the target drone and the spinning camera induces significant scale changes in MSR.
To prevent regression degradation for distant, small-scale targets, we replace this baseline with Efficient IoU (EIoU) loss~\cite{zhang2022focal}:
% To avoid limiting the regression accuracy for distant, small-scale targets, we replace the baseline with Efficient IoU (EIoU) loss ~\cite{zhang2022focal}.
\begin{equation}
    \begin{aligned}
        \mathcal{L}_{EIoU} = \mathcal{L}_{IoU} + \frac{\rho^2(b, b_{gt})}{c^2} + \frac{\rho^2(w, w_{gt})}{w_c^2} + \frac{\rho^2(h, h_{gt})}{h_c^2},
    \end{aligned}
\end{equation}
where $\rho(\cdot)$ represents the Euclidean distance, $c$ is the diagonal length of the smallest enclosing box covering the predicted box $b$ and the ground truth box $b_{gt}$ , while $w_c$ and $h_c$ denote its width and height, respectively.  
By independently regularizing the width and height to provide direct gradients, EIoU ensures consistent localization accuracy and robust bounding box regression across a wide range of operational distances, from close-range encounters to long-distance tracking.

\subsubsection{Bearing Estimation}
\label{sec:bearing_estimation}

\begin{figure}[t]
    \centering
    \includegraphics[width=1\linewidth]{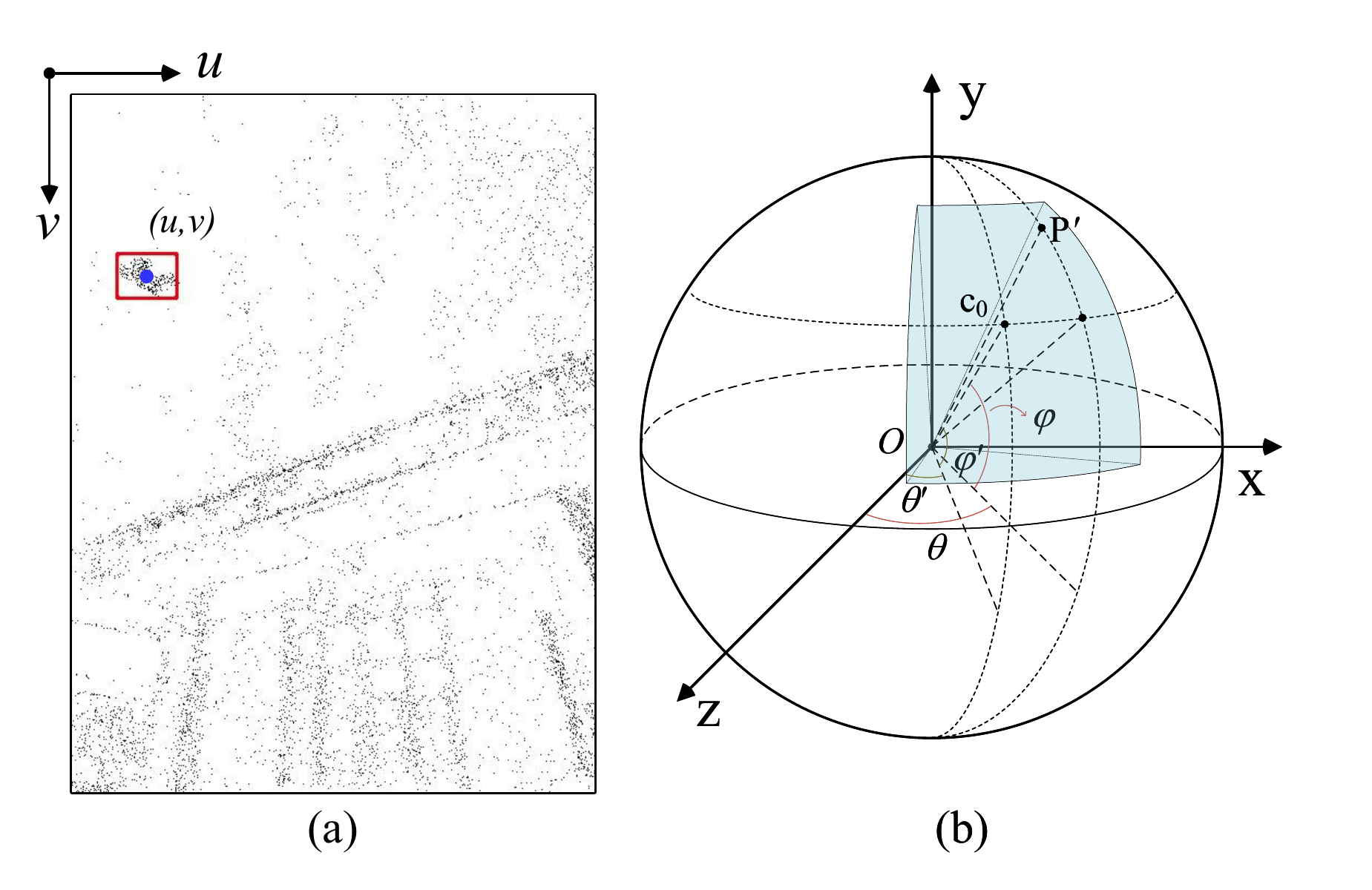}
    \caption{
    \textit{Coordinate systems and bearing estimation.}
    % 太长了 两个坐标系的示意图 Detection image
    (a) Image plane: Detection result with a red bounding box marking the drone and a dot indicating its centroid.
    (b) Spinning platform system: The detected centroid is projected into the 3D coordinate system (origin: camera optical center; y-axis: rotation axis; z-axis: zero-direction). This projection enables bearing calculation.}
    \label{fig: bearing_estimation}
    \vspace{-2em}
\end{figure}
% 图注:无人机坐标位置由image平面到云台坐标系的转换示意图 (a)为检测用的图片，红框为无人机位置，点是我们认为的无人机的中心 (b)为(a)投影到以相机光心为原点，旋转平台的转轴为y轴，旋转平台零点角度方向为z轴

% \zs{The bearing estimation follows the projection method used in CMax-SLAM~\cite{guo2024cmax}. 
% Once a drone is detected, its bearing relative to the platform is computed for subsequent interaction or navigation. }
After obtaining the detection results from the network, we compute the bearing to the drone relative to our system using a projection method inspired by CMax-SLAM~\cite{guo2024cmax}.
As illustrated in Fig.~\ref{fig: bearing_estimation} (a), the drone centroid $(u,v)$ is defined as the center of the detection bounding box.

The normalized bearing vector in image coordinates is
\begin{equation}
    \mathbf{X}_\mathrm{P} = \mathbf{K}^{-1}\left(u,v,1\right)^{\top},
    \label{eq:bearing_vector}
\end{equation}
where $\mathbf{K}$ is the camera intrinsic matrix. This vector is then transformed into the spinning platform's coordinate system using the rotation:
\begin{equation}
    \mathbf{X}_{\mathrm{P}^{\prime}} = \mathbf{R}(\theta')\mathbf{R}(\phi')\mathbf{X}_\mathrm{P},
    \label{eq:bearing_rotation}
\end{equation}
where $\mathbf{R}(\theta')$ and $\mathbf{R}(\phi')$ represent the rotation matrices for the spinning platform’s motion and the camera’s fixed installation tilt, respectively.
The instantaneous angle of the spinning platform is given by $\theta' = (t - t_{0})\omega$, where $t_{0}$ is the reference time when the platform passes the zero direction, $t$ is the detection timestamp, and $\omega$ is the angular velocity.

Finally, the rotated point $\mathbf{X}_{\mathrm{P}'} = (X_{\mathrm{P}'}, Y_{\mathrm{P}'}, Z_{\mathrm{P}'})^{\top}$ (see Fig.~\ref{fig: bearing_estimation}(b)) is transformed into the spherical coordinate system, with its angular representation given by
\begin{equation}
    \begin{aligned}
    \phi = \arcsin \frac{Y_{\mathrm{P}'} }{\sqrt{X_{\mathrm{P}'}^2 + Y_{\mathrm{P}'}^2 + Z_{\mathrm{P}'}^2}}, \quad 
    \theta = \arctan \frac{X_{\mathrm{P}'}}{Z_{\mathrm{P}'}},
    \end{aligned}
    \label{eq:spherical_coords}
\end{equation}
where $\phi$ and $\theta$ represent the estimated bearing of the drone in the spinning platform coordinate system.

\section{Experiments}
\label{sec: experiments}

This section presents a comprehensive evaluation of the proposed system to validate its effectiveness and efficiency.
% We construct a dedicated dataset using our system because existing datasets are incompatible with our setup, and state-of-the-art detection algorithms fail in our target scenario, making direct comparison with prior work infeasible.
Our evaluation is structured as follows: 
We first introduce the experimental details (Sec.~\ref{subsec: experimental_details}).
Then, we analyze the performance of the network in lighting conditions and present an ablation study (Sec.~\ref{subsec: detection_network}).
Finally, we demonstrate the real-world performance through outdoor onboard tests (Sec.~\ref{subsec:outdoor_experiments}).

\subsection{Experimental Details}
\label{subsec: experimental_details}

\subsubsection{Training Setup}
The network is trained on a server equipped with an NVIDIA RTX A6000 GPU (48 GB VRAM), a 30-core Intel Xeon Platinum 8336C CPU, and 60 GB of RAM.
% We take the YOLOX framwork as the backbone.
A batch size of 64 is used with the Adam optimizer at an initial learning rate of 0.0001. 
The loss function combines EIoU loss, L1 loss, classification loss, and binary cross-entropy loss. 
% binary cross-entropy loss
All layers are randomly initialized from scratch.

\subsubsection{Experimental Setup}
The proposed system is mounted on a Unitree Go2-W quadruped robot to enable autonomous locomotion. 
% Onboard computation is performed on an NVIDIA Jetson Orin NX with 16 GB LPDDR5 memory, operating in the 25 W power mode with 8 CPU cores and 4 GPU TPCs enabled.
Onboard computation is handled by an NVIDIA Jetson Orin NX (16 GB LPDDR5) configured for maximum performance (25 W mode).
The model is optimized and converted to TensorRT with FP16 precision.
A DJI Mini 4K serves as the detection target.

\subsubsection{Ground Truth Acquisition}
Ground truth trajectories and poses are obtained by fusing data from GNSS receivers on both our device and the drone, together with a MID-360 LiDAR integrated with the quadruped robot.
The poses of our device are further refined using FAST-LIO~\cite{xu2021fast}, while trajectory fitting and GPS–LiDAR alignment generate the carrier’s trajectory and pose at each time instant.
To minimize timing errors, our system adopts strict time synchronization, with a particular design worth highlighting: an STM32 microcontroller processes both the PPS signal from the GNSS module and the trigger from the photoelectric switch before forwarding them to the camera. As the camera accepts only a single external trigger, the two signals are configured at different frequencies to ensure reliable separation.
%zs:逻辑改为”为了减少timing errors，我们的系统进行了严格的时间同步，其中特别值得介绍的是：......“

\subsubsection{Baseline Considerations}
% Direct quantitative comparison with existing state-of-the-art (SOTA) event-based detectors is not feasible in our specific operational context.
% Current SOTA models are predominantly designed for static viewpoints or low-speed ego-motion, heavily relying on standard spatial accumulation or recurrent memory that assumes stable background contexts.
% Under our continuous spinning scenario, the severe spatial misalignment and continuous background flow cause these standard representations to fail entirely, resulting in near-zero recall.
% Consequently, rather than comparing against incompatible architectures that cannot process uncompensated rotational event streams, we establish our spatial-only detection backbone (the model without TFM) as the primary baseline.
% This rigorously isolates and validates the contribution of our proposed spatio-temporal framework under high-dynamic conditions.
Current SOTA event-based methods rely on specific assumptions, such as static backgrounds~\cite{da2025new} or low-speed motion~\cite{chen2025eMoE-Tracker}, to maintain stable representations.
These priors are violated by the severe spatial misalignment in our spinning scenario, making these methods impractical on our data.
% Instead, we propose a more generalized framework that leverages the high-temporal-resolution nature of events without external stability constraints.
Instead, our framework is explicitly designed for continuous high-dynamic ego-motion, leveraging the high temporal resolution nature of events without relying on stable background priors.
Consequently, we employ our spatial-only backbone as the primary baseline to rigorously isolate the TFM’s contribution under high-dynamic conditions.

\begin{figure}[t]
    \vspace{1.0em}
    \centering
    \includegraphics[width=1\linewidth]{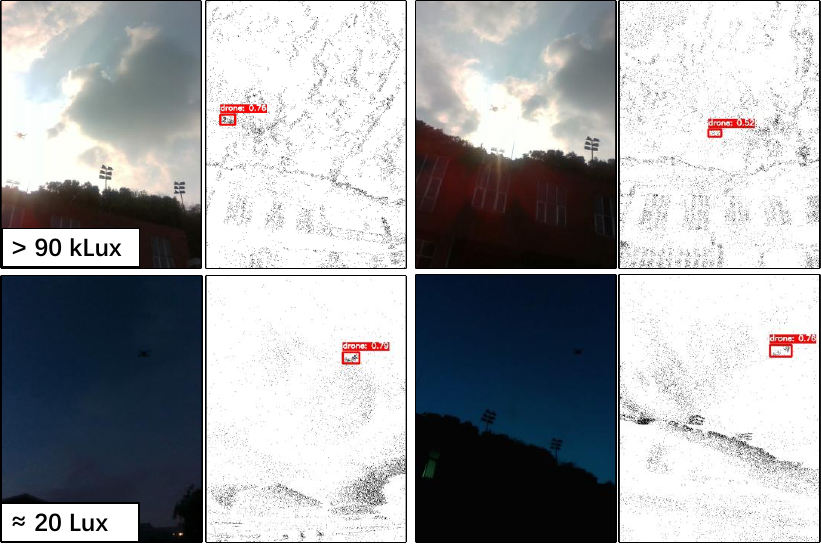}
    \caption{
    \textit{Illustration of detection results under varying lighting conditions.} 
    Top: High-glare scenario with the sun directly within the camera's FoV during the afternoon. 
    Bottom: Low-light scenario after sunset.}
    \label{fig: illumination}
    \vspace{-0em}
\end{figure}

\begin{table}[t]
\vspace{-0em}
\begin{center}
\caption{\textit{Evaluation under varying lighting conditions} [\%].
        % Summary of key detection metrics under three lighting conditions. 
        Minimum Detectable Size (MDS) denotes the smallest target width identifiable by the network under the corresponding scenario.
        In high-glare scenarios, we deem that size-based comparison is not particularly meaningful; therefore, this aspect is not included in the experimental evaluation.
}
\vspace{0em}
\setlength{\tabcolsep}{3mm}{
\begin{tabular}{l c c c c}
\hline
\toprule
\textbf{Illumination}   & \textbf{AP}  & \textbf{AP\textsubscript{75}} & \textbf{AR\textsubscript{100}}  & \textbf{MDS [pixels]} \\    % Minimum Detectable Size (pixels)
\midrule
Standard                & 62.19                                                   & 69.70                 & 67.12                              & 14                                           \\
Low                     & 59.30                                                   & 63.80                 & 62.60                              & 17                                           \\
High                    & 55.80                                                   & 58.20                 & 60.20                              & -                                           \\
\bottomrule
\hline
\end{tabular}
}
\label{tab:illumination}
\vspace{-3em}
\end{center}
\end{table}

\subsubsection{Evaluation Metric}

The drone's bearing in the world coordinate system is represented as a unit vector derived from its spherical coordinates $(\theta, \phi)$, where $\theta$ and $\phi$ denote the azimuth and elevation angles, respectively.
To quantify the estimation accuracy, we compute the angular error $\gamma$ between the estimated bearing vector $\vec{v}_{\text{est}}$ and the ground truth vector $\vec{v}_{\text{GT}}$. This error is defined as the true 3D angle between the two vectors:
\begin{equation}
    \gamma = \arccos(\vec{v}_{\text{est}} \cdot \vec{v}_{\text{GT}}).
    \label{eq:angular_error}
\end{equation}
This metric accurately measures the shortest angular distance between the two directions in 3D space.

For the detection performance, for the performance of the network, we adopt Average Precision (AP), AP at 75\% IoU (AP\textsubscript{75}), and Average Recall at 100 detections (AR\textsubscript{100}) as the main evaluation metrics for recognition results.

\subsection{Detection Performance}
\label{subsec: detection_network}

In this section, we evaluate the performance of the network, beginning with an assessment of key detection metrics under challenging illumination conditions, followed by an ablation study to validate the contribution of the proposed module.
% This specialized evaluation is necessary as existing datasets are incompatible with our representation method, and state-of-the-art detection algorithms fail in our target scenario, making conventional benchmarking against prior work infeasible.
\begin{table}[t]
\vspace{1.0em}
\begin{center}
\caption{\textit{Ablation study of temporal fusion.}
Up or down arrows ($\uparrow$ or $\downarrow$) indicate whether a higher or lower value is better.
Network parameters (Params) represent the total number of learnable weights in the network.}
\vspace{0em}
\setlength{\tabcolsep}{3mm}{
\begin{tabular}{ccc}
\hline
\toprule 
\textbf{Metric} & \textbf{w/o Temporal Fusion} & \textbf{w/ Temporal Fusion} \\
\midrule 
\textbf{AP$\uparrow$ [\%]} & 62.19 & \textbf{64.92} (\scriptsize{+2.73}) \\
\textbf{AP\textsubscript{75}$\uparrow$ [\%]} & 69.70 & \textbf{72.80} (\scriptsize{+3.10}) \\
\textbf{AR\textsubscript{100}$\uparrow$ [\%]} & 67.12 & \textbf{69.87} (\scriptsize{+2.75}) \\
\textbf{AP\textsubscript{S}$\uparrow$ [\%]} & 56.30 & \textbf{59.90} (\scriptsize{+3.60}) \\
\textbf{Params $\downarrow$ [M]} & \textbf{8.94} & 9.53 \\
\bottomrule
\hline
\end{tabular}
}
\label{tab:ablation}
\vspace{-4em}
\end{center}
\end{table}

\subsubsection{Performance Under Varying Lighting Conditions}
To validate the robustness of the proposed network in outdoor conditions, we evaluate its performance under three selected lighting scenarios: standard illumination, low-light conditions, and high-glare conditions with the sun within the camera's FoV.

Fig.~\ref{fig: illumination} provides a visualization of these lighting conditions. 
Under high illumination (top row), solar over-exposure highlights clouds while suppressing intensity changes from the drone and structures, causing the drone to appear washed out and blend into the background. 
Under low light (bottom row), reduced contrast and lack of self-emission make it difficult to detect the drone without propeller motion; lower event counts introduce more noise, further complicating detection.
Quantitative results (Tab.~\ref{tab:illumination}) show strong performance in standard lighting. 
Under strong illumination, metrics decline due to feature loss from overexposure, though baseline detection remains feasible. In low light, accuracy is maintained with only slight degradation.

\begin{figure*}[t]
\vspace*{1mm}
    \centering
    \includegraphics[width=1\linewidth]{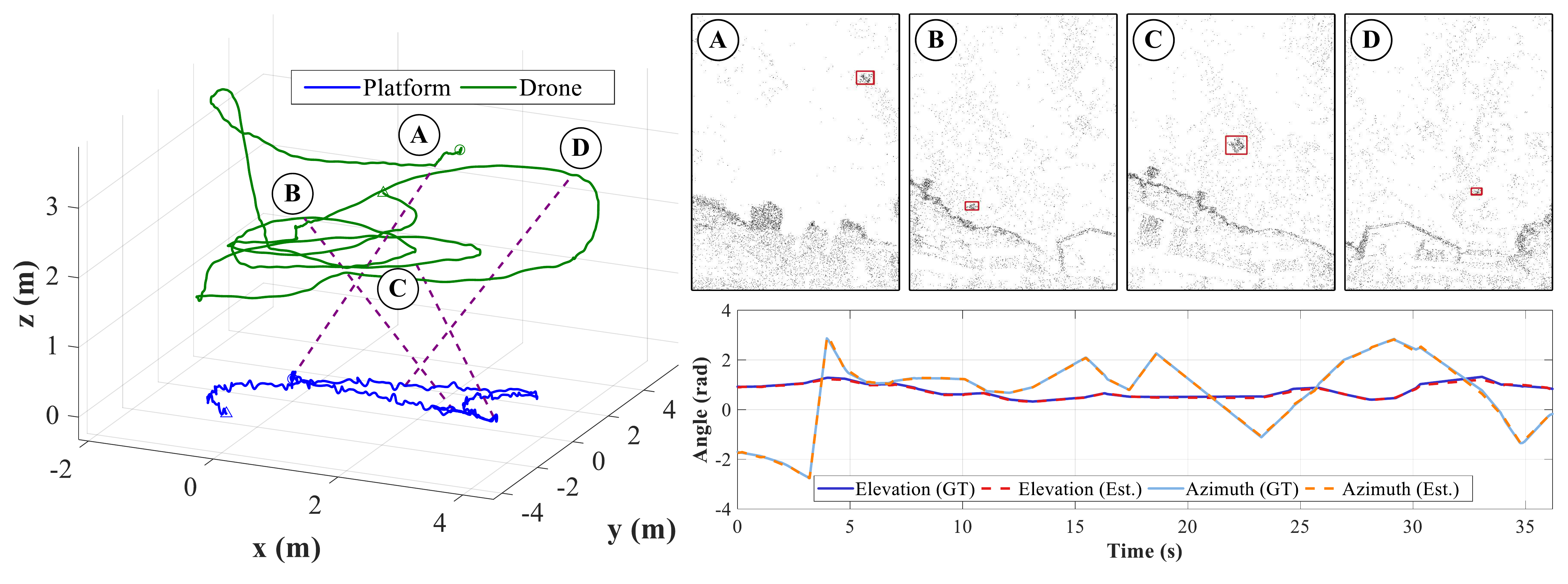}
    \caption{
    \textit{Illustration of the detection results.} 
    Left: Trajectories of the carrier and drone, with four representative time instants marked by circled letters and connected by purple dashed lines. 
    Top right: Corresponding camera images showing the detection results, where drone detection results are highlighted. 
    Bottom right: Comparison of the estimated angles (Est.) with the ground truth (GT), with GPS data interpolated using cubic splines to obtain the drone’s position at the exact detection moments.
}
\label{fig: results}
\vspace{-1em}
\end{figure*}

% 文字/线要加粗，图注，文字统一 GT
% 左图是平台和无人机的运动轨迹，选择其中四个点单独标出，中间是对应左图中标记的四个点的相机的图像，右边是整个过程中误差，横轴是误差，纵轴是处于对应误差范围的点的出现次数
In summary, these experiments confirm that our event-based detection system is not only effective under ideal conditions but also exhibits superior robustness to extreme illumination variations that typically hinder frame-based approaches.
This capability is a direct result of the inherent properties of the event sensor, combined with our designed processing pipeline, underscoring the system's suitability for real-world deployment where lighting cannot be controlled.

\subsubsection{Ablation Study}

In this section, we validate the impact of the Temporal Fusion Module, the key contribution of the system, on the performance of the network.
The results of the ablation study are summarized in Tab.~\ref{tab:ablation}.
Consistent improvements are observed across all major detection metrics when the Temporal Fusion Module is incorporated, notably in AP at multiple thresholds and recall rates. 
The parameter count increases slightly.
However, subsequent onboard experiments further confirm that this modest rise in parameter count does not compromise the real-time performance of the system, demonstrating that the proposed module enhances representational capacity efficiently without incurring significant computational overhead.

\begin{table}[t]
\vspace{-0em}
\begin{centering}
\caption{Evaluation on self-recorded sequences using angular error [°].
}
\vspace{0em}
\setlength{\tabcolsep}{2.5mm}{
\begin{tabular}{c c c c}
\hline
\toprule
\textbf{Sequences}   & \textbf{Mean error[°]}  & \textbf{Median error[°]} & \textbf{Max error[°]}  \\
\midrule
\textit{static\_1}         &2.063    &2.096        &3.103   \\
\textit{static\_2}         &1.772    &1.764        &2.611   \\
\textit{static\_3}         &1.736    &1.715        &2.835   \\
\textit{static\_4}         &1.858    &1.848        &3.144   \\
\textit{moving\_1}         &2.004    &1.857        &4.340   \\
\textit{moving\_2}         &1.927    &1.921        &4.308   \\

\bottomrule
\hline
\end{tabular}
}
\label{tab:error}
\vspace{-3em}
\end{centering}
\end{table}

\subsection{Outdoor Experiment}
\label{subsec:outdoor_experiments}

To assess the practicality of our system, we conduct six outdoor experiments, comprising four sequences under stationary conditions and two under carrier motion. 
The evaluation results are presented in Tab.~\ref{tab:error}.
The stationary sequences yield stable and concentrated error distributions, whereas the moving sequences exhibit more dispersed errors with larger peaks, reflecting the impact of dynamic disturbances. 
It should also be noted that the ground truth data in the dynamic experiments rely on the Fast-LIO algorithm, which may degrade under rapid motion; hence, the reported errors may stem from both system inaccuracies and reference uncertainties.

The results indicate that azimuth ($\theta$) estimation remains accurate, whereas elevation ($\phi$) exhibits larger deviations.
This is likely because locomotion-induced jitter of the quadruped robot has a greater impact on pose interpolation accuracy in the vertical direction.
Quantitatively, sequence \textit{moving\_1} yields a mean error of about 2°, with deviations occasionally exceeding 4°, confirming reliable performance under motion-induced fluctuations.
Despite the increased error under motion, the system maintains functional performance and provides usable bearing estimates even under the challenging conditions of quadruped locomotion.
We visualize the bearing estimation results on a moving carrier in Fig.~\ref{fig: results}.
Despite the severe ego-motion, the estimates closely align with the ground truth, demonstrating the system's capacity to maintain robust and reliable target localization
% The operational details are visualized in Fig.~\ref{fig: results}, depicting the experimental trajectories, detection snapshots, and the corresponding bearing estimates. As shown, the estimated curves closely track the ground truth throughout the sequence, providing a clear demonstration of the system's system's high estimation accuracy.

% The practical detection performance on a moving carrier is illustrated in Fig.~\ref{fig: results}. 
% The figure visualizes the trajectories of both the carrier and the drone, highlighting detection snapshots at representative timestamps (A-D). 
% The bearing estimates effectively track the ground truth trajectory, demonstrating the system's capacity to maintain continuous and reliable target localization despite the severe ego-motion of the carrier.
\begin{table}[t]
\vspace*{3mm}
\centering
\caption{Computational performance on different platforms [$\mathbf{Time}$: ms].}
\label{tab:time_consume}
\vspace{0.5em}
% \begin{adjustbox}{max width=\textwidth}
\renewcommand\arraystretch{1.2}
\setlength{\tabcolsep}{3mm}
\begin{tabular}{c l r r}
\toprule
\textbf{Node} & \textbf{Function} & \textbf{Laptop} & \textbf{Jetson NX} \\
\midrule
\multirow{2}{*}{\makecell[c]{Pre-processing}} 
    & Data processing & 0.969 & 1.938 \\
    & Image generation & 10.841 & 30.514 \\
\midrule

\multirow{5}{*}{\makecell[c]{Detection}} 
    & Input construction & 3.762 & 5.223 \\
    & Inference & 11.440 & 20.425 \\
    & Bearing estimation & 0.0475 & 0.0823 \\
    & Others & 15.249 & 18.467 \\
\addlinespace[0.2em]
\cdashline{2-4}
\addlinespace[0.2em]

& Subtotal & 27.335 & 44.938 \\
\bottomrule
\end{tabular}
% \end{adjustbox}
\vspace{-2.5em}
\end{table}

\subsection{Computational Efficiency}

% To assess the computational performance, we evaluate our algorithm on two different platforms: a laptop (AMD Ryzen R7 5800H, RTX 3060 GPU), and computing platform embedded in the system (NVIDIA Jetson Orin NX).
To assess the computational performance of our algorithm, we conduct experiments on the \textit{moving\_1} sequence using two different platforms: a laptop (AMD Ryzen R7 5800H, RTX 3060 GPU) and an embedded platform (NVIDIA Jetson Orin NX).
% Both platforms are evaluated on the \textit{moving\_1} sequence to ensure consistency across experiments. 
Table~\ref{tab:time_consume} reports the average runtime of each processing module,  providing a detailed breakdown of the computational resource allocation on different hardware.

Our system, implemented in C++, consists of two processes.
The pre-processing node converts the events into image, with the entire process executed on the CPU.
The computational bottleneck is the detection node, which involves \textit{input construction}, \textit{inference}, and \textit{bearing estimation}. 
The Inference part is accelerated with TensorRT, while the other components are processed on the CPU. 
The "Others" category primarily includes the time required for transferring event representations from the CPU to the GPU.
The subtotal runtime of the detection node confirms that our system can operate stably in real time at 30 Hz on the desktop platform and approximately 22 Hz on the embedded platform.
% The implementation, developed in C++, consists of two nodes: a pre-processing node and a detection node. 
% Latency is mainly distributed across three stages: event-to-image generation, neural network inference, and input construction for network preparation. 
% Among them, image generation and inference dominate the overall runtime on both platforms. 
% The pipeline runs in real time, achieving 30 Hz on the laptop and 22 Hz on the Jetson Orin NX with multi-threaded execution.
% These results demonstrate that our method strikes a favorable balance between accuracy and efficiency, making it suitable for real-time drone detection in resource-constrained environments.

% The comparable latency among these stages, each constituting roughly one-third of the total time, highlights a balanced computational pipeline and underscores the importance of co-designing sensing, computation, and data handling for embedded aerial perception.

\section{Conclusion}
\label{sec:conclusion}
% \dai{We present ODD-SEC, a real-time drone detection and bearing estimation system using a spinning event camera for 360° horizontal perception on moving platforms.}
% \joey{We present ODD-SEC, a system that enables 360° horizontal perception for moving platforms by using a spinning event camera to detect drones and estimate their bearing in real time.}
% A temporally-aware YOLOX-based network processes multi-frame event representations, extracting spatial and temporal features to handle motion blur and massive data volume caused by camera's ego-motion.
% The system estimates drone bearing in azimuth and elevation and runs in real-time on a Jetson Orin NX. 
% Outdoor experiments demonstrate robust performance under stationary and dynamic conditions, achieving a mean angular error of 1.91°. 
% Future work will explore trajectory prediction, autonomous navigation and mapping, and closed-loop deployment in complex environments, alongside code and dataset release to support further research.

We present ODD-SEC, a system that enables 360° horizontal perception on moving carriers to detect drones and estimate their bearing in real time.
By mounting an event camera on a spinning platform, our system significantly expands the FoV, overcoming the inherent FoV limitations of traditional event camera setups.
Correspondingly, we propose a novel event representation and a lightweight neural network, which allow drone detection without motion compensation and further realize bearing estimation, even when the carrier is in motion.
Extensive outdoor experiments under both stationary and dynamic conditions validate the accuracy and robustness of our approach, yielding a mean angular error of only 1.91°, and demonstrate real-time performance at 22 Hz on Jetson Orin NX.
Future work will explore trajectory prediction, autonomous navigation and mapping, and closed-loop deployment in complex environments, along with code release to support further research.

%我们提出了ODD-SEC，一个能够在移动平台上实时全向检测无人机的系统。
%通过将事件相机固定在旋转平台上，我们的系统极大的扩展了FoV，克服了the inherent FoV limitations of traditional event camera setups.
%相应的，我们提出了一种新的事件表征形式与对应的轻量级网络，即使平台运动也能在不进行运动补偿的情况下 
%广泛的室外实验既证明了我们算法的准确性与鲁棒性，无论平台静止还是运动（此处可以列举具体数字）；也展示了我们系统能够在Jeston.....上实时运行。
%我们会......

\bibliographystyle{IEEEtran}
\bibliography{myBib}
\end{document}